\documentclass{article}
\usepackage[utf8]{inputenc}
\usepackage{amsmath, amssymb}
\usepackage{hyperref}
\usepackage{graphicx}
\usepackage{float}
\usepackage[T1]{fontenc}
\usepackage[french]{babel}
\usepackage{csquotes}
\usepackage[utf8]{inputenc}
\usepackage[table]{xcolor}
\usepackage{caption}
\usepackage{array}        
\usepackage{multirow}     
\usepackage{colortbl}     
\usepackage{array}
\usepackage{etoolbox}
\usepackage{authblk}
\usepackage{bbm}

\AtBeginEnvironment{tabular}{\centering}
\setkeys{Gin}{keepaspectratio=true}
\usepackage{enumitem}
\setlist[itemize]{label=\textbullet}

\usepackage[T1]{fontenc} 
\usepackage{lmodern} 
\usepackage{microtype} 

\usepackage[shortcuts]{extdash} 
\usepackage{hyphenat} 

\addto\captionsfrench{}
\addto\captionsfrench{\renewcommand{\abstractname}{Abstract}}
\title{\textbf{Polysemy of Synthetic Neurons \\  
\large Towards a New Type of Explanatory Categorical Vector Spaces}}

\author{Michael Pichat$^{1,2,4}$, William Pogrund$^{1,5}$, Paloma Pichat$^{1,3}$, Judicael Poumay$^{1}$, Armanouche Gasparian$^{1}$, Samuel Demarchi$^{1,4}$, Martin Corbet$^{1,2}$, Alois Georgeon$^{1,2}$, Michael Veillet-Guillem}

\date{}

\affil[1]{Neocognition (Chrysippe R\&D)}
\affil[2]{Facultés Libres de Philosophie et de Psychologie de Paris (ER IPC)}
\affil[3]{Faculté de Médecine de Lyon Est (Université Lyon 1)}
\affil[4]{Université Paris 8}
\affil[5]{INP-PHELMA, Université Grenoble Alpes}
\begin{document}

\maketitle

\renewcommand{\abstractname}{Abstract}
\begin{abstract}

The poly-semantic nature of synthetic neurons in artificial intelligence language models is currently understood as the result of a necessary superposition of distributed features within the latent space. We propose an alternative approach, geometrically defining a neuron in layer \( n \) as a categorical vector space, with a non-orthogonal basis, composed of categorical sub-dimensions extracted from preceding neurons in layer \( n-1 \). This categorical vector space is structured by the activation space of each neuron and enables, via an intra-neuronal attention process, the identification and utilization of a critical categorical zone for the efficiency of the language model—more homogeneous and at the "intersection" of these different categorical sub-dimensions.

\end{abstract}

\section{Theoretical Context}

\subsection{Neuronal Activations and the Structure of the Vector Space and Semantic Encoding}

The organization of semantic representations in large language models (LLMs) relies on a specific structuring of the vector space, shaped by neuronal activations. These activations, originating from Feed-Forward Network (FFN) blocks, do not merely transmit signals through the network: they actively contribute to the construction and organization of the semantic concepts learned by the neural network \cite{Wang2024, Han2025}.

The contribution of each neuron to the structuring of the vector space depends on the intensity of its activation, which is generally proportional to the absolute value of its activation and the L2 norm of the corresponding vector \cite{Wang2024}. Positive and negative activations segment the vector space into distinct regions: the former are directly incorporated into the residual stream and project their values towards the model's vocabulary, while the latter influence the weighting of semantic representations differently \cite{Wang2024}.

This structuring is subject to internal regularization constraints, particularly Magnitude Consistency, according to which activations within the same layer tend to exhibit a homogeneous norm \cite{Pham2024}. This property suggests that activations should not be understood as independent points in space but rather as oriented vectors. The activation space at the individual neuron level can thus be conceptualized as a (d-1)-dimensional hypersphere, where directional adjustments allow modifications to semantic representations without altering the overall stability of the vector space \cite{Pham2024}. Empirical findings support the idea that the encoding of semantic concepts is based on this structured organization of activations.

The targeted manipulation of neuronal activations allows direct control over the semantic entropy of the model's generations: by bounding certain activations between specific values (clamping), it is possible to directly modulate the probability distribution of generated tokens, thereby steering the model's output toward semantically determined content \cite{Han2025}. Advanced techniques, such as Householder Pseudo-Rotation (HPR), facilitate these adjustments by modifying the orientation of neuronal activations while preserving their norm, ensuring a coherent transformation of the vector space \cite{Pham2024}.

This structuring extends to multilingual models, where neuronal activations vary depending on the language of the processed text. Multilingual LLMs exhibit specific linguistic regions, where distinct sets of neurons are activated based on the input language. The targeted deactivation of these regions results in a significant degradation of performance in the corresponding language while leaving others unaffected, suggesting an effective segmentation of the vector space by language \cite{Zhang2024}. Simultaneously, these models align the representations of similar semantic concepts across languages within a common latent space (\textit{Lingua Franca}), facilitating the transfer of linguistic capabilities \cite{Zhang2024}.

This organization evolves with model training and scaling. As the model grows in size, linguistic regions tend to fragment, while semantic alignment strengthens, indicating a convergence towards a more unified representation of concepts  \cite{Zhang2024}. A similar structuring is observed in multimodal models (MLLMs), where neuronal activations specialize according to the input modality. Neuron attribution methods, such as NAM (Neuron Attribution Method), allow the identification of neuron subsets dedicated to textual inputs (T-neurons) and visual inputs (I-neurons), revealing a clear separation of activations by modality \cite{Fang2025}.

Neuronal activations also influence the contextual coherence of LLMs' generations. Certain neurons dynamically adjust their activations to maintain semantic stability within a given context, as evidenced by a significant correlation between activation coherence and the semantic complexity of inputs \cite{Vitiello2024}. Moreover, these activations reflect semantic proximity between words more accurately than simple statistical co-occurrence. For instance, synonyms produce more similar neuronal activations than words that merely share a common context of occurrence, highlighting the role of activations in the semantic organization of representations \cite{Digutsch2023}.

All these observations suggest that neuronal activations constitute structuring functional units in the organization of the vector space and the encoding of semantic concepts in LLMs.

\subsection{Polysemy of Neurons and Superposition}

A series of studies postulate that deep language models exploit a superposition dynamic to encode a greater number of features than the available dimensions in these models. This phenomenon relies on the use of distributed and quasi-orthogonal representations in high-dimensional latent spaces, enabling efficient information compression \cite{Hanni2024}. One of the notable effects of this organization is the emergence of poly-semantic neurons, i.e., neurons activated by conceptually distinct inputs.

Contrary to the intuitive idea of a one-to-one correspondence between a neuron and a feature, language models do not assign a specific characteristic to a single neuron. Instead, they represent them in a distributed manner within a linear subspace of the layer’s residual stream rather than in a single neuron \cite{Nanda2023}. These features can be extracted through linear projections, revealing that information is both compressed and spread across multiple neurons.

This superposition is believed to result from a structural constraint. Models must store more information than they have neurons, which forces them to organize their activations compactly \cite{Bills2023}. One observed mechanism is the use of antipodal pairs, where a single neuron encodes two opposing concepts to maximize activation differentiation \cite{Hanni2024}.

A consequence of this superposition is the emergence of poly-semantic neurons, empirically observed in the MLP layers of transformers, where some neurons respond simultaneously to highly different inputs. A single neuron may be activated by both images of dogs and the concept of democracy, complicating its interpretation \cite{Elhage2022}.

Brickens et al. (2023) \cite{Bricken2023} suggest that this polysemy is not arbitrary but results from the shared use of neuronal resources and the distributed structuring of activations. Rather than strict specialization, each neuron is designed to encode multiple features simultaneously, generating a semantically complex distribution of activations. This organization follows precise geometric principles, where representations emerge from the linear combination of multiple feature directions.

Mathematically, superposition relies on structuring the activation vector space to minimize interference between encoded representations. An adaptation of the Johnson-Lindenstrauss lemma to neural networks suggests that it is possible to encode a large number of features \( k \) in a space of dimension \( d \), with \( k \gg d \), as long as the dot product between their representations remains low:
\[
\| W_i \cdot W_j \| \ll 1, \quad \text{for } i \neq j
\]
This constraint promotes exponential information compression by exploiting optimized geometric structures, such as polytopes, to organize representations \cite{Hanni2024}.

Another key factor is the impact of critical non-linearities, particularly ReLU activation functions. These play a filtering role by reducing interference noise between superimposed features. They impose response thresholds, stabilizing representations and improving model robustness \cite{Adler2024}.

Several studies document the impact of superposition at different levels of representation. At the lexical level, some neurons respond simultaneously to heterogeneous token categories, demonstrating partial or even complete overlap in their activations \cite{Hoang2024}. At a more abstract level, certain neural units encode semantically distant concepts simultaneously, revealing an adaptive sharing of activation space \cite{Bills2023}.

The learning dynamics of models also favor a progressive interweaving of representations. Backpropagation and self-organization jointly optimize multiple tasks, enabling the simultaneous use of a single neuron for different functions depending on the input context \cite{Zhuang2024}. Even locally specialized neurons can contribute to multiple cognitive processes within a dynamic and adaptive framework \cite{AlKhamissi2024a}.

In their 2025 study, Haider et al. \cite{Haider2025} demonstrate, however, that poly-semantic neurons do not activate multiple concepts simultaneously in an undifferentiated manner but rather encode them within specific activation value ranges, forming distinct Gaussian distributions with minimal overlap. To reach this conclusion, they analyzed neuronal activations in both encoder and decoder language models, studying their behavior across multiple text classification datasets. Using a statistical and qualitative approach, they showed that neuron activations follow distinct patterns for each concept, allowing each activation range to be associated with a specific concept.

\subsection{Superposition and Quasi-Orthogonality}

Orthogonality refers to an arrangement where vectors encoding different knowledge or concepts are perpendicular in the latent space, thereby minimizing interference \cite{Mikolov2013}. Such an organization promotes a clear separation of information and reduces the effects of catastrophic forgetting, where new knowledge disrupts previously acquired knowledge \cite{Ramsauer2021}. However, perfect orthogonality is rarely achieved, as it would imply infinite storage capacity, which is unattainable in a constrained latent space \cite{Liu2023}. In practice, models balance an approximately orthogonal organization with a certain degree of redundancy, optimizing both compression and flexibility in representations \cite{Gao2024}.

Superposition interacts with the phenomenon of quasi-orthogonality, defined as a state where neuronal representations, while close to orthogonality, are not perfectly perpendicular in latent space \cite{Bricken2023}. In LLMs, this quasi-orthogonal state limits interference between knowledge or tasks represented in distinct directions \cite{Gao2024}. However, this orthogonality remains partial: a perfectly orthogonal encoding would eliminate all interference, but models must contend with residual overlaps that increase as more knowledge is added \cite{Touvron2023}. 

Quasi-orthogonality is closely linked to the concept of superposition, as it serves as an optimization strategy within a necessarily limited representational space \cite{Cunningham2023}. The ability to maintain an approximately orthogonal encoding facilitates the poly-semantic compression of numerous features while preserving generalization capacity, as previously mentioned. However, more pronounced interference can ultimately degrade overall performance \cite{Touvron2023}.

From a knowledge management perspective, quasi-orthogonality allows for the addition or modification of new information without significantly disrupting existing knowledge \cite{Meta2024a}. When orthogonality is nearly perfect, conflicts between knowledge representations remain minimal. However, as orthogonality deviates from this ideal, interference accumulates, increasing the risk of degradation or catastrophic forgetting \cite{Wang2023a}. In this dynamic, the emergence of parameter collisions presents a major challenge: multiple tasks end up sharing the same subspaces within the model, reducing effective orthogonality and amplifying interference \cite{Wang2023a,Meta2024b}. Various techniques have been proposed to mitigate these collisions, including N-LoRA methods, which attempt to preserve or restore orthogonality between subspaces associated with each training instance, thereby limiting the unintended rewriting of previously acquired knowledge \cite{Wang2023a}.

Large-scale language models with greater parametric capacity generally exhibit an increased tendency toward orthogonality \cite{Touvron2023,Meta2024b}. Their expanded representational space allows for a finer separation of relevant dimensions, reducing interference noise. Consequently, this finer-grained separation of concepts and tasks translates into superior performance, particularly in complex tasks requiring intensive knowledge management \cite{Touvron2023}. However, larger models also face novel challenges in detecting and correcting internal collisions, as the multiplicity of parameters and potentially addressed tasks makes controlling residual orthogonality more difficult \cite{Meta2024b}.

A deeper understanding of superposition in LLMs, whether manifested through quasi-orthogonality or parameter collisions, largely depends on disentanglement techniques. Concept Activation Vectors (CAVs) enable the projection of neural activity into semantically interpretable directions \cite{Nicolson2024}. Sparse Autoencoders (SAEs) offer a dimensionality reduction method to isolate semantic features into distinct latent codes \cite{Cunningham2023}, making residual overlaps visible. Furthermore, causal intervention approaches, which involve selectively manipulating the activation of certain neurons, have demonstrated the importance of overlaps in determining model outputs: any targeted modification can simultaneously affect multiple tasks or concepts. For instance, \cite{Yamakoshi2023} applied causal interventions to transformer models to analyze pronoun resolution, showing that targeted manipulation of specific attention heads influences the propagation of contextual information and simultaneously affects multiple possible interpretations. Similarly, \cite{He2023} studied causal relationships across different layers of language models, revealing that specific interventions modulate neural activity hierarchically and affect various cognitive functions in parallel \cite{Yamakoshi2023,He2023}.

This superposition has significant implications for the explainability of large language models (LLMs) and the localization of their internal knowledge. Probing methods indicate that neurons initially appearing specialized and monosemantic (e.g., concerning grammatical categories) actually participate in more abstract functions when considering the entire context. For instance, \cite{Gurnee2023} demonstrated that the early layers of LLMs use sparse neuron combinations to represent multiple superimposed features, while intermediate layers contain neurons that appear less poly-semantic and more dedicated to high-level contextual characteristics. \cite{Zhao2023} suggests that this situation complicates the traceability of predictions, as multiple neurons share competing semantic loads. At the same time, superposition combined with partial quasi-orthogonality may enhance robustness and generalization, as a shared set of heavily trained neuronal resources supports multiple linguistic facets \cite{Bills2023}.

Finally, on a practical level, the existence of poly-semantic units and quasi-orthogonal regions opens possibilities for new forms of selective fine-tuning. \cite{Xu2025} show that specifically targeting certain neurons or subspaces to improve a given task can limit interference with other functions. \cite{Xiong2024} also report that LLMs frequently encapsulate multiple tasks within shared subspaces, suggesting that controlled manipulation of these subspaces could facilitate model adaptation to new domains. The parallel drawn by \cite{Wang2024} between these mechanisms and neural plasticity in the human brain suggests that superposition, combined with partial quasi-orthogonality, is not merely a technical artifact but a systemic organizational principle essential to the functioning and rapid learning of large language models.

\subsection{Dimensional Representations of Concepts in Human Cognition}

What about these questions regarding the relationship between activation and semantic encoding in humans? Semantic concepts in the human brain are encoded by distributed neural patterns that, similarly to vectors used in artificial intelligence, reproduce the structure of meaning \cite{Franch2025, Zhang2020, Sassenhagen2020}. Cortical regions, particularly the anterior ventral temporal lobe, organize this information into graded, multidimensional spaces where vector proximity reflects semantic similarity between concepts \cite{Cox2024, Zhang2020, Lonshakov2024}, aligning with findings on graded semantic spaces obtained through Representational Similarity Learning methods \cite{Cox2024}. The hippocampus plays a crucial role in the vector encoding of meaning by integrating context into a dynamic and distributed coding system, akin to mechanisms observed in large language models \cite{Franch2025, Chen2025, Desbordes2023}.

The organization of concepts in the brain follows a spatial and multidimensional logic, where semantic similarity is reflected in the proximity of neural activations within cortical networks. In this perspective, the duality between similarity and cognitive representation constitutes a fundamental principle of mental organization. The notion of psychological space, as described by \cite{Roads2024}, highlights how cognitive structuring relies on perceived distances between concepts, thereby encoding their semantic relationships. Various formal models, including geometric, set-theoretic, and graph-based representations, capture these mental structures based on cognitive tasks and stimuli. The study of these psychological spaces is grounded in a second-order isomorphism, where the internal structure of representations must preserve the relationships between external stimuli rather than their intrinsic properties.

The temporal lobe, particularly in its anterior ventral region as noted, structures concepts in a graded manner by translating semantic similarity into distances within an internal vector space, allowing for a fine organization of representations \cite{Cox2024, Karlgren2021, Zhang2020}. Diffuse cortical networks distribute semantic representations through overlapping activation patterns, revealing a multidimensional organization relying on decoding schemes derived from neural activity \cite{Zhang2020, Nishida2021, Wang2023}. Context integration results in a progressive increase in dimensionality as a sentence unfolds, highlighting that semantic complexity evolves dynamically and is reflected in the increasing neural signals \cite{Desbordes2023, Palenik2024, Chen2025}.

Neural activations, particularly in the inferior frontal gyrus, exhibit a geometry comparable to that of contextual embeddings in large language models, illustrating a geometric alignment between the brain and AI \cite{Goldstein2024, Franch2025, Zhang2020}. Semantic distances measured in hippocampal activity positively correlate with representations from contextual models such as BERT, whereas they diverge from non-contextual models like Word2Vec, underscoring the importance of context in semantic encoding \cite{Franch2025, Sassenhagen2020, Nishida2021}. Furthermore, the alignment of multimodal representations, obtained through fMRI data analysis, demonstrates that models integrating both visual and textual modalities best replicate brain activity in the associative cortex \cite{Nakagi2024, Goldstein2024, Lonshakov2024}.

The study by Stoewer (2022) \cite{Stoewer2022} further supports this perspective by highlighting the role of cognitive maps in structuring abstract concepts. The hippocampus and entorhinal cortex are not limited to spatial navigation but actively contribute to the formation and organization of cognitive maps of semantic space. By applying multiscale successor representation, their neural model organizes concepts based on their perceived similarity, forming hierarchical structures where fine-grained conceptual relations are preserved in detailed representations, while higher levels yield broader categories. These representations also enable efficient interpolation of new or incomplete concepts, suggesting that human cognition leverages predictive mechanisms similar to those used in artificial intelligence to process complex semantic spaces.

The semantic complexity of a stimulus induces an increase in the dimensionality of neural representations, indicating an adaptive coding system that integrates multiple aspects of meaning \cite{Desbordes2023, Chen2025, Sheng2022}. Hybrid symbolic and distributed coding manifests through a complementarity between categorical organization and information distribution, thus merging symbolic and connectionist approaches in concept processing \cite{Nakagi2024, Karlgren2021, Zhang2020}. The high dimensionality of neural representations is a performance indicator for episodic memory and fine language comprehension, as confirmed by studies linking these dimensions to robust behavioral correlates \cite{Chen2025, Palenik2024, Nishida2021}.

Word embeddings illustrate the ability of computational models to reproduce the geometric representation of meaning by aligning vector structures with empirically observed neural activation patterns \cite{Karlgren2021, Sassenhagen2020, Goldstein2024}. Language processing models demonstrate an aptitude for predicting cortical responses across various brain regions, further validating the hypothesis of shared semantic encoding between the brain and AI \cite{Zhang2020, Nishida2021, Franch2025}.

Artificial language models replicate the neural structuring of meaning. This thus supports the idea of a strong analogy between categorical segmentation mechanisms of reality (concept formation) in vivo and in silico.

\subsection{Synthesis of Our Previous Work: Activation / Categorization Homomorphism \& Extraction of Categorical Sub-Dimensions}

After reviewing a series of recent studies in both artificial intelligence and human cognition concerning the relationship between neural activation and semantic encoding, we now present a synthesis of our work in this field. Specifically, we focus on the complex relationship, within artificial neural networks, that can be identified between activation and synthetic neural categorization.

More precisely, our approach consists of questioning the following: how do synthetic neurons in language models create "thought categories" to segment and analyze their informational environment, their "techno-umwelt" \cite{Efimov2023}? And what is the role of neural activation in this framework? What are the categorical and activation-related characteristics, at the level of formal neurons, of this artificial cognition? How should we interpret the modality and function of the apparent "polysemy" of this categorical cognition? Our investigation of these questions falls within a neuropsychological approach to the explainability of artificial language neural networks. This approach consists of using concepts from cognitive psychology and human cognitive sciences as relevant heuristic landmarks to construct interpretative frameworks for synthetic cognition, which, while avoiding the pitfall of anthropocentrism, remain aligned with human cognitive modalities.

From both a theoretical and methodological perspective, our specific approach involves studying these questions through the relationship, for given tokens, between their activation levels and their categorical proximity, that is, the link between activation and "semantics." This is done to understand the extent to which, for a given neuron, the segmentation of its activation space carries epistemological and operational significance in terms of categorical segmentation. More precisely, within a genetic cognitive perspective, we seek to understand how activation impacts the process of extracting categorical sub-dimensions from the thought categories of layer \( n-1 \), in order to form the new, more functional thought categories that constitute the neurons of layer \( n \).

Our previous exploratory studies on these topics have exclusively focused on the perceptron-type layers 0 and 1 of OpenAI's GPT2-XL language model. More specifically, they have examined the core tokens of the involved neurons, defined as the 100 tokens that activate each neuron most strongly on average, using data from \cite{Bills2023}. Regarding the measurement of categorical proximity between tokens, we used the observation framework provided by the input embeddings of GPT2-XL, via cosine similarity calculations.

In our first empirical study, we highlighted the following phenomena of synthetic neural cognition related to the question of neural polysemy:  
\begin{itemize}
    \item \textbf{Categorical discontinuity} \cite{Pichat2024a}: For a given neuron, within its segment of moderately strong activations, successive core tokens—based on their activation level—are not particularly close in categorical terms and may even be relatively distant.
    \item \textbf{Categorical inhomogeneity} \cite{Pichat2024a}: Within the segment of moderately strong activations, core tokens with the same activation level are not necessarily categorically close and may be relatively distant.
    \item We grouped these discontinuity and inhomogeneity phenomena under the synthetic process of \textbf{categorical divergence (peri-activational)} \cite{Pichat2024a}, within the segment of moderately strong activations.
\end{itemize}

However, we later identified a phenomenon that opposes neural polysemy: \textbf{categorical convergence} \cite{Pichat2024b}. That is, as the activation levels of successive core tokens increase, the categorical proximity between these core tokens also increases.

These categorical divergence and convergence phenomena align with the observations of \cite{Bricken2023}, who noted that neurons appear monosemantic when examined with their top-activated tokens but reveal themselves to be poly-semantic when studied based on tokens with lower activation levels.

We interpreted these seemingly paradoxical results as follows: the categorical dimension constructed by a neuron in layer \( n \) (or, more precisely, by its aggregation function, among other factors) can be decomposed into a vector space of categorical sub-dimensions \cite{Pichat2024b}. These sub-dimensions result from a projection—by mathematical construction of the neuron's aggregation function—of the categorical output dimensions of its precursor neurons in layer \( n-1 \). In this framework, core tokens with moderately strong activations belong to a single categorical sub-dimension, and their categorical divergence arises from the fact that two activationally close tokens may trigger different categorical sub-dimensions. Conversely, core tokens with very high activations simultaneously engage multiple categorical sub-dimensions. This accounts for their very high activation levels—again by mathematical construction of the aggregation function—and their categorical convergence results from the reduction of categorical degrees of freedom generated by such an "intersection" of categorical sub-dimensions.

We have previously postulated that the categorical dimension carried by a "poly-semantic" target neuron (in layer \( n \)) can be interpreted as a vector space of categorical sub-dimensions. These sub-dimensions are extracted from the categorical output dimensions of its associated precursor neurons in the preceding layer \( n-1 \). How are these categorical sub-dimensions abstracted, and what are their cognitive characteristics, particularly regarding the question of polysemy?

The algebraic operations performed by the neural aggregation function (of the form \( \sum (w_{i,j} x_{i,j}) + a \)) serve as key epistemological and operational markers for identifying the mathematical-cognitive factors that genetically shape this categorical sub-dimension extraction \cite{Pichat2024c}. These factors act as operators in the categorical reconstruction of the informational world performed by artificial cognition. Within this framework, we have identified three fundamental processes governing the extraction of categorical sub-dimensions:

\begin{itemize}
    \item \textbf{Categorical priming} \cite{Pichat2024c}, or \textbf{the \( x \) effect}: The more a token is activated within a precursor neuron in layer \( n-1 \), the more likely it is to be activated (and thus extracted) within a target neuron (strongly connected) in layer \( n \), thereby forming the categorical extension of the latter.
    
    \item \textbf{Inter-neuronal categorical attention} \cite{Pichat2024c}, or \textbf{the \( w \) effect}: The stronger the connection weight of a target neuron (layer \( n \)) to a precursor neuron (layer \( n-1 \)), the more a token that is strongly activated in the precursor neuron tends to be activated (and thus extracted) within the target neuron, thereby contributing to its categorical extension. This synthetic phenomenon qualitatively manifests as a process of \textit{categorical complementation}: the core tokens involved, at the level of the target neuron, belong to categorically different sub-dimensions, each associated with inter-neuronal categorical attention to a different precursor neuron. This categorical complementation explains the previously mentioned phenomenon of categorical divergence, and thus polysemy.
    
    \item \textbf{Categorical phasing} \cite{Pichat2024c}, or \textbf{the  \( \sum \) effect}: The more a token is (strongly) activated across multiple neurons in layer \( n-1 \), the more likely it is to be activated (and thus extracted) within a strongly connected target neuron in layer \( n \), thereby contributing to its categorical extension. This artificial cognition process qualitatively manifests as a phenomenon of \textit{categorical intersection}: the core tokens involved, at the level of the target neuron, simultaneously belong to multiple categorical sub-dimensions, which resonate in a form of "categorical echo." This categorical intersection accounts for the previously highlighted phenomenon of categorical convergence, thereby limiting the poly-semantic dimension of neurons. As indicated, this categorical convergence occurs for tokens with very high activation levels, where such extreme activations are directly generated by the aggregation function through its mathematical construction.
\end{itemize}

These three mathematical-cognitive factors, as we have just seen, guide the selective extraction of specific tokens at the level of layer \( n-1 \) neurons, with these particular tokens generating the extension (the content) of each categorical sub-dimension thus extracted within a neuron in layer \( n \).

Categorical priming, inter-neuronal categorical attention, and categorical phasing are the causal architectural and mathematical factors governing the genetic, relatively poly-semantic extraction of categorical sub-dimensions during the transition from layer \( n \) to layer \( n+1 \). How, then, can we qualitatively understand the categorical abstraction process that emerges from this? And through what quantitative modalities does this constrained poly-semantic abstraction operate in a functional manner?

Le détourage catégoriel (Pichat et al., 2024d) est le processus par lequel la cognition synthétique extrait, au niveau d’un neurone d’arrivée en couche n, une sous-dimension catégorielle de la dimension catégorielle (i.e. de la catégorie de pensée) de sortie de chacun de ses neurones précurseurs associés, en couche n-1, avec lequel il est fortement lié. Ce phénomène d’abstraction permet, au niveau de chaque neurone précurseur, la sélection d’une forme catégorielle originale (dont l’extension est un ensemble de tokens déterminés) à « détacher » d’un fond catégoriel. Cette segmentation catégorielle découpe et fabrique ainsi, dans le monde des tokens, de nouvelles sous-dimensions catégorielles ; chacune de ces dernières étant portée par un sous-ensemble de tokens catégoriellement homogènes et alignés avec la nouvelle catégorie spécifique, polysémique de façon mesurée, que crée et porte leur neurone d’arrivée correspondant ; l’union géométrique de ces sous-dimensions générant alors la nouvelle dimension catégorielle associée au neurone d’arrivée impliqué. 

Categorical clipping \cite{Pichat2024d} is the process through which synthetic cognition extracts, at the level of a target neuron in layer \( n \), a categorical sub-dimension from the categorical dimension (i.e., the artificial thought category) of the output of each of its strongly connected precursor neurons in layer \( n-1 \). This abstraction phenomenon allows, at the level of each precursor neuron, the selection of an original categorical form (whose extension consists of a specific set of tokens) to be "detached" from a categorical background. This categorical segmentation thus isolates and constructs, within the world of tokens, new categorical sub-dimensions. Each of these sub-dimensions is supported by a subset of categorically homogeneous tokens, aligned with the newly created specific category, which remains moderately poly-semantic, and is carried by its corresponding target neuron. The geometric union of these sub-dimensions then generates the new categorical dimension associated with the target neuron involved.  

We have identified several functional characteristics of this categorical clipping:

\begin{itemize}
    \item \textbf{Categorical reduction} \cite{Pichat2024d}, referring to the fact that the extracted categorical sub-dimension is associated with a set of (taken-)tokens \footnote{We define a "taken-token" as a core-token of a neuron in layer n-1 that is also a core-token of its associated neuron in layer n. A categorical sub-dimension extracted by a neuron in layer n from a neuron in layer n-1 is therefore, by definition, composed of taken-tokens in terms of its extension.}
 that are more categorically homogeneous among themselves compared to the entire (core-)token set of the precursor category. This limits polysemy within a given categorical sub-dimension.

    \item \textbf{Categorical selectivity} \cite{Pichat2024d}, which manifests as an extraction of a quantitatively more restricted extension in terms of the number of (taken-)tokens, relative to the precursor neuron's category.

    \item \textbf{Separation of initial embedding dimensions} \cite{Pichat2024d}, indicating that categorical clipping, when observed in the vector space of the initial embeddings of GPT2-XL, tends to manifest as an elective dichotomous compartmentalization of these embeddings. Some embeddings become preferentially associated with the extracted categorical form (i.e., the sub-dimension), while others remain more aligned with the categorical background "left behind" in the original category.

    \item \textbf{Segmentation of categorical zones} \cite{Pichat2024d}, visible through a relative displacement of the categorical gravity centers of both the extracted form and the non-retained background. Each of these barycenters is positioned in contrasting categorical regions.
\end{itemize}

These various properties help us understand the synthetic and operational modalities through which categorical clipping creates extractions of categorical sub-dimensions by grouping tokens that converge within a fabricated homogeneous categorical segment. These sub-dimensions are constitutive of neuronal polysemy.

Categorical restructuring \cite{Pichat2025} is the process through which each new neural layer clips (abstracts) and combines relevant categorical sub-dimensions from the artificial thought categories of its previous layer. This process aims to shape new, more efficient categories for analyzing and processing the unique experience that the synthetic system acquires from the linguistic external world to which it is exposed. What are the cognitive characteristics of this categorical restructuring? Within this restructuring, is the clipping of a categorical sub-dimension associated with a specific segmentation at the activation space level of the involved source or target neurons?We have highlighted various properties of synthetic categorical restructuring related to the issue of neuronal polysemy:

\begin{itemize}
    \item \textbf{Partial categorical confluence} \cite{Pichat2025}: This characteristic of relative polysemy arises from the fact that the categorical sub-dimensions clipped at the level of a target neuron in layer \( n \), from the categories of its predecessor neurons in layer \( n-1 \), tend to partially converge in a semantic space, contributing to the formation of the neuron's relatively poly-semantic category. This phenomenon is particularly related to the coactivity of inter-neuronal attention and categorical phasing, which together generate the extraction of tokens belonging to multiple categorical sub-dimensions.

    \item \textbf{Categorical activation dispersion} \cite{Pichat2025}: A categorical sub-dimension extracted from the category of a neuron in layer \( n-1 \) does not correspond to a continuous segment of activation values for the relevant tokens within the activation space of either the source or target neuron. This means that activation segments neither delimit nor define the relatively poly-semantic categorical sub-dimensions that will be clipped. Indeed, categorical priming and inter-neuronal attention alone are insufficient to clip a categorical sub-dimension—they require the additional effect of categorical phasing.

    \item \textbf{Categorical distancing} \cite{Pichat2025}: The artificial thought categories carried by neurons in layer \( n \) are semantically significantly different from those carried by their strongly connected predecessor neurons in layer \( n-1 \). This is immediately generated by the categorical uniqueness of the categorical sub-dimensions clipped from layer \( n-1 \) to \( n \). Categorical distancing is, of course, the primary function of successive synthetic neural layers, gradually compensating for the initial categorical limitations of the model's starting embeddings.
\end{itemize}

Our \textit{\href{https://neuron-viewer.neocognition.org/1/5}{neurone viewer génétique}} allows visualization of the properties we have just described regarding the phenomenon of synthetic categorical restructuring occurring in the transition from perceptron layer 0 to 1 in GPT2-XL.

Through the phenomenon of categorical activation dispersion \cite{Pichat2025}, we have seen that, for a given source or target neuron, its activation space segments do not delimit the clipped (relatively poly-semantic) categorical sub-dimensions. In what other possible neuronal topology might activation segmentation be linked to categorical segmentation? And to what extent could this potential relationship be connected to the extraction of categorical sub-dimensions that remain relatively poly-semantic?

In investigating this question, we have highlighted the synthetic phenomenon of \textbf{intra-neuronal attention} \cite{Pichat2025a}, which refers to a neuron's ability to detect and isolate a homogeneous categorical segment within the poly-semantic thought category it carries, based on a segmentation of its activation space. This phenomenon, subtle but systematic, shows that formal neurons can establish a homomorphic relationship between activational and categorical segmentations—but only at the level of the highest-activated tokens.

We have identified the following properties of this intra-neuronal attention:

\begin{itemize}
    \item \textbf{Activational indifferentiation} \cite{Pichat2025a}: A minimal difference in mean activation values between categorical clusters (categorical subgroups constituting a neuron's poly-semantic category) associated with lower mean token activation values.

    \item \textbf{Relative activational differentiation} \cite{Pichat2025a}: A greater difference in mean activation values between poly-semantic categorical clusters involving one or more higher mean token activation values.

    \item \textbf{Activational entanglement of categorical clusters} \cite{Pichat2025a}: The poly-semantic categorical clusters of a neuron are not compartmentalized into distinct activational segments but rather overlap across these segments.

    \item \textbf{Low categorical homogeneity of activational segments} (in which a neuron's activation space can be segmented), associated with a relative increase in categorical homogeneity \cite{Pichat2025a} (and thus a reduction in polysemy) as the mean activation value of these activational segments increases.
\end{itemize}

Thus, activation would play the role of an intra-neuronal attentional vector: it would allow a neuron to detect and delineate, among the full set of tokens that constitute its poly-semantic categorical extension, those tokens specifically associated with the highest activation values—tokens that are relatively more homogeneous from a categorical standpoint (i.e., less poly-semantic), in line with the previously highlighted phenomenon of categorical convergence.

An artificial neuron can be interpreted as a synthetic cognitive agent of conceptualization (in the sense of Vergnaud\cite{Vergnaud2009, Vergnaud2016}), whose purpose is to attentively select, from a set of poly-semantic tokens to which it reacts strongly, a specific subset of tokens. This subset constitutes the categorical extension of the "critical" concept-in-action, which is less poly-semantic and which the neuron is functionally designed to detect selectively. The neuron thereby performs an attentional focusing activity on specific types of tokens that must be selected and filtered to enable, in a subsequent step, the clipping of a categorical sub-dimension from this critical concept-in-action. The paroxysmal concept-in-action thus determined consists of tokens with very high activation levels and lower polysemy (at least concerning the conceptual characteristic being constructed). These particular tokens tend to converge semantically (categorical homogeneity, low polysemy). The relative activational differentiation of this critical synthetic concept-in-action is closely tied to categorical priming (effect \( x \)), as it inherently denotes tokens with very high activation. Effect \( x \), combined with categorical phasing and inter-neuronal categorical attention, then enables the clipping of a categorical sub-dimension. Conceptualization and intra-neuronal attention, both operating at the level of a layer \( n \) neuron, are thus the fundamental synthetic processes that subsequently allow, at the next level of layer \( n+1 \), the formation of new thought categories, even more functional for executing the tasks assigned to a neural network.

Conceptualization, again in the sense of Vergnaud \cite{Vergnaud2009, Vergnaud2016}, is the fundamental activity through which cognition identifies the only relevant characteristics of the objects (in this case, poly-semantic tokens) on which it must operate—properties whose consideration is central to the efficiency of the cognitive system's action. Conceptualization is described as "in action" by Vergnaud because it is neither consciously processed, justified, nor verbalized by the cognitive system involved; rather, it is progressively recognized as functional relative to the tasks to be performed. This conceptualization leads to the formation of concepts-in-action, which are poly-semantic categories of thought capturing the essential information being sought (here, specific properties of tokens, to be grouped within the category associated with a neuron). Theorems-in-action, on the other hand, are rules indicating how these concepts-in-action should be combined to create new ones that are more refined and specific (less poly-semantic). 

In this framework, the aggregation function, again of the form \( \sum (w_{i,j} x_{i,j}) + a \), can be interpreted as a theorem-in-action specifying, at the level of a given target neuron in layer \( n \), how to specifically combine the concepts-in-action formed at the level of precursor neurons in layer \( n-1 \) to construct the new concept-in-action specific to this target neuron. At the level of a layer \( n-1 \) neuron, and in its paroxysmal form—when effects \( x \), \( w \), and \( \sum \) reach their peak (in relation to a layer \( n-2 \) neuron)—this theorem-in-action enables the determination of a critical concept-in-action, which is less poly-semantic, for this \( n-1 \) neuron. From this critical concept-in-action, the clipping of a relatively monosemous categorical sub-dimension (at least regarding the conceptual characteristic constructed at this level) is then performed at the level of the \( n \) neuron.

\section{Problematic}

Our previous research suggests the existence, albeit subtle yet systematic, of a relationship between activation segmentation and categorical segmentation (i.e., polysemy reduction) for a given neuron in the early layers of GPT2-XL, specifically at the level of its tokens with very high activation magnitudes. We have postulated that, within a given neuron in layer \( n \), intra-neuronal attention allows it to identify and isolate, from a set of poly-semantic tokens, a critical concept-in-action, which constitutes a subset of the category associated with this neuron. This critical concept-in-action, exhibiting stronger categorical convergence and thus reduced polysemy, corresponds to very high activation values of the tokens it involves. This results from the effect of categorical phasing (operating in layer \( n-1 \)), forming a categorical zone at the intersection of various categorical sub-dimensions that geometrically constitute the category of this neuron—sub-dimensions clipped from the neural thought categories in layer \( n-1 \). These very high activation values are unique to this paroxysmal and less poly-semantic concept-in-action, whereas other activation segments of the same neuron have lower activation values. This is because the tokens involved at these lower levels each belong to only one categorical sub-dimension, rather than to their intersection, which leads to polysemy reduction. At the level of neuron \( n+1 \), a new, less poly-semantic categorical sub-dimension will in turn be clipped from this critical concept-in-action in layer \( n \), particularly through the effect of categorical priming made possible by the strong activation associated with this critical concept-in-action. This relatively monosemous categorical sub-dimension, along with other sub-dimensions clipped from layer \( n \), will in turn geometrically constitute the poly-semantic vector space (of categorical sub-dimensions) characteristic of each neuron in layer \( n+1 \).

Our previous postulates suggest that the poly-semantic thought category associated with a neuron in an early layer \( n \) can be geometrically interpreted and represented as a categorical vector space, whose basis consists of the categorical sub-dimensions (relatively monosemous) clipped by this neuron from layer \( n-1 \)—a non-orthogonal basis due to the previously mentioned partial categorical confluence. This categorical vector space is structured by the activation space of the neuron as follows: within this categorical vector space, tokens with very high activation values simultaneously belong to multiple categorical sub-dimensions, generating categorical convergence and thus a reduction in their polysemy. In contrast, tokens with lower activation values, being more poly-semantic, each belong to only one categorical dimension at a time, generating categorical divergence among these tokens.

Various studies, aside from our own, addressing the question of neuronal polysemy, describe it as a necessary factor for encoding a number of neural thought categories greater than the number of neurons available in an MLP layer. According to this perspective, synthetic categories are encoded through an activation pattern involving multiple neurons. Conversely, these neurons can be activated in relation to a wide variety of different thought categories, thereby enabling the encoding of a very large number of synthetic categories. This mechanism contributes to the performance of large language models via the principle of superposition \cite{Elhage2022, Bricken2023, Nanda2023}.

These studies thus interpret neuronal polysemy as a result of a synthetic thought category being encoded through a distribution of activations within a vector space composed of the activations of the entire (or a subset of the) neurons in a given layer. As an alternative, but not contradictory, approach to these studies, we propose another type of vector space that could be invoked to explain and model neuronal polysemy—or at least the semantic phenomenology that appears poly-semantic to us based on our own human categorical observation framework. This vector space is not located outside a given neuron (i.e., across an inter-neuronal set of neurons within a layer) but rather inside the neuron itself. It is an intra-neuronal vector space, whose basis is composed of categorical sub-dimensions (clipped from neurons in the immediately preceding layer) that constitute this neuron. Our approach to polysemy thus reverses the relationship between categories and neurons: whereas other studies conceptualize a poly-semantic synthetic category as being distributed across different neurons, our perspective suggests that a poly-semantic neuron itself contains a variety of categorical sub-dimensions, which are precisely the source of its polysemy. It is important to clarify, once again, that our interpretation is not opposed to the other perspectives we have discussed but rather presents an alternative viewpoint. Indeed, just as in physics an elementary "particle" can manifest as either corpuscular or wave-like depending on the observational methodology used, the phenomenon of polysemy—like any phenomenon—can be expressed or manifested differently within various interpretative observational frameworks.

More operationally, and for the purpose of testing our epistemologically alternative approach, we hypothesize the following: for a given neuron, highly activated tokens, in contrast to those with lower activation, are the ones that simultaneously exhibit a higher level of categorical proximity (i.e., dimensional proximity) to the categorical sub-dimensions (of the categorical vector space) of this neuron. More broadly, we propose the hypothesis of a functional, monotonic, and positive relationship between token activation levels and their coordinates within the vector space formed by the categorical sub-dimensions of the involved neurons.

\section{Methodology}

\subsection{Methodological Framework}

To methodologically position our study, we provide an overview of the main explainability approaches applied to artificial neural networks. These techniques aim, at various levels of analysis, to elucidate the internal semantic dynamics of models by explaining either the underlying mechanisms of information flow or the relationship between the network’s inputs and outputs, whether studied at the scale of individual layers, groups of layers, or the model as a whole.

The so-called “macroscopic” explainability methods examine variations between input data and model predictions to highlight correlations between what is fed into the synthetic system and what it produces. Among these approaches, gradient-based methods quantify the influence of each input dimension by analyzing the partial derivatives of outputs with respect to inputs \cite{Enguehard2023}. Input interpretation can also be performed through the extraction of distinctive features \cite{Danilevsky2020}, the assessment of the importance of elementary units (tokens) \cite{Enguehard2023}, or the analysis of attention coefficients \cite{Barkan2021}. Another class of example-based methods observes output variations in response to specific modifications of inputs, whether minor perturbations \cite{Wang2022} or more substantial alterations such as deletion, negation, rearrangement, or occlusion of input elements \cite{Atanasova2020, Wu2020, Treviso2023}. Other studies adopt a mapping approach, linking inputs to outputs through conceptual representations \cite{Captum2022}.

The so-called “microscopic” explainability methods, on the other hand, focus on the internal intermediate states of models rather than on input-output relationships alone. They analyze neuronal activations or interactions between different network units to better understand their functional role. Some studies explore the propagation and transformation of activations across the model’s layers \cite{Voita2021}, while others propose modifications to activation functions to enhance their readability and interpretability \cite{Wang2022}. Additional approaches aim to extract knowledge encoded in neural networks by projecting internal representations through interpretation matrices \cite{Dar2023, Geva2023}. Finally, certain research efforts rely on statistical analysis of neural responses to specific datasets to identify recurring structures and draw conclusions about the model’s cognitive organization \cite{Bills2023, Mousi2023, Durrani2022, Wang2022, Dai2022}. It is within this latter line of inquiry that our current study is situated.

\subsection{Model and Statistical Units Used}

Following our previous research \cite{Pichat2024a, Pichat2024b, Pichat2024c, Pichat2024d, Pichat2025, Pichat2025a}, our study is based on data from the GPT2-XL model. This choice aligns with our ongoing exploration of concept formation within large language models. It was made using data provided by OpenAI in 2023, in connection with the study conducted by Bills and collaborators that same year, as well as the valuable insights associated with it.

Our analyses specifically focused on the second MLP layer of GPT2-XL, encompassing 6,400 neurons and the 100 most highly activated tokens on average for each of these neurons—an ensemble we refer to as "core-tokens." Each neuron in layer L1 was associated with 0 to 10 taken-clusters, where a taken-cluster is a group of core-tokens shared between the L1 neuron and one of the ten L0 neurons with the highest connection weights to it.

\subsection{Statistical Processing}

Our statistical analyses focus on two primary variables for each neuron: the activation rate of its core-tokens and their dimensional proximity score with each of the involved categorical subdimensions.

In general, we compare groups of tokens based on these proximity scores and their activation levels within the neuron in question. Given that these analyses rely on a limited number of observations, we prioritize non-parametric tests, which are better suited for small sample sizes and the absence of normality assumptions.

Accordingly, we apply the following statistical tests depending on the context:
\begin{itemize}
    \item The Kruskal-Wallis test, used to compare multiple groups, followed by an appropriate \textit{post hoc} test in case of significant differences.
    \item The Wilcoxon-Mann-Whitney test, employed for the comparison of two groups.
    \item The $\chi^2$ test, used to examine the distribution of results at the network scale.
    \item Kendall's tau, allowing for the evaluation of nonlinear correlation between dimensional proximity scores and activations.
\end{itemize}

\section{Results}

\subsection{Relationship Between Activation and Dimensional Proximity: Group Comparisons}

As a reminder, our goal is to assess whether there is a relationship between the activation level of a neuron for a given token and the number of principal categorical vector subdimensions to which this token is linked (dimensional proximity). 

From a methodological perspective, our approach was structured as follows:
\begin{itemize}
    \item For each target neuron in layer 1 of GPT2-XL, we focused on its 10 primary taken-clusters. These clusters are defined as the 10 subgroups of tokens that are common to (i) its core-tokens (i.e., the 100 tokens that activate this target neuron the most on average) and (ii) the core-tokens of its 10 most strongly connected precursor neurons from layer 0. Each of these 10 taken-clusters constitutes what we define as the extension of a principal categorical subdimension of the category associated with this target neuron. Note that for a given neuron, the number of taken-clusters may be fewer than 10 if it does not meet all the intersection criteria mentioned above. However, this does not affect the calculations described below (i.e., the averaging is performed over $n<10$ subdimensions rather than 10). For each of the 100 core-tokens $j$ of a target neuron $i$, and for each principal categorical subdimension $k$ of this neuron, we computed the dimensional proximity score $s_{ijk}$. This score was operationalized as the mean cosine similarity (via GPT2-XL embeddings) between the core-token $j$ and all tokens forming the taken-cluster associated with categorical subdimension $k$. Hence, $s_{ijk}$ can be interpreted as the coordinate of core-token $j$ on the principal subdimension $k$, or as the extent to which this core-token $j$ categorically belongs to this subdimension $k$.
    
    \item We then computed, for a given neuron, the average $s_{ij}$ of the dimensional proximity scores of token $j$ across all 10 principal categorical subdimensions $k$. This value $s_{ij}$ thus expresses the extent to which token $j$ has jointly high coordinates across the 10 categorical subdimensions or the average coordinate of this token $j$ across the 10 categorical subdimensions $k$ associated with the given neuron.
\end{itemize}

Table n°1 presents the relationship between token activation levels and the average dimensional proximity of these tokens to the (up to) 10 principal categorical subdimensions of each involved neuron in layer 1 of GPT2-XL. To ensure statistical robustness, we retained only neurons ($n=614$) that met the following criteria: (i) associated with at least three taken-clusters, each containing at least six taken-tokens, and (ii) having a cumulative total of at least 40 different tokens across their taken-clusters. Furthermore, for each target neuron in layer 1, we selected its 10 least activated (min) and most activated (max) core-tokens and then computed their corresponding mean dimensional proximity scores $s_{min}$ and $s_{max}$. We observe that the percentage of neurons for which the average difference in dimensional proximity, $d = s_{max} - s_{min}$, is positive is high (78.01\%); this result is highly significant at the inferential level ($P_{\chi}(d>0)<.0001$) based on a $\chi^2$ goodness-of-fit test under an equiprobability distribution hypothesis. The observed average difference in dimensional proximity, $d = .0274$, is not large but remains highly meaningful. It corresponds to a Cliff's delta effect size of $D_c = .2407$, which, while moderate, is non-negligible. This effect size is particularly noteworthy given that, by definition, core-tokens exhibit relatively dense categorical homogeneity, which naturally limits large differences in dimensional proximity. Notably, a comparison involving $s_{min}$ derived from less activated tokens outside the core-token set could potentially yield much stronger differences, further supporting our hypothesis. This likely explains why, at the local level of each neuron, inferential differences (measured using a Kruskal-Wallis approach) remain relatively modest (\% of ($P_{KW}(d > 0) < .05$) = 22.8\%).

\begin{table}[H]
    \centering
    \renewcommand{\arraystretch}{1.3}
    \begin{tabular}{|c|c|}
        \hline
        $n$ & 614 \\
        \hline
        Mean $\alpha_{\text{min}}$ & 1.3843 \\
        Mean $\alpha_{\text{max}}$ & 3.0218 \\
        \hline
        Mean $\sigma_{\text{min}}$ & .3991 \\
        Mean $\sigma_{\text{max}}$ & .4265 \\
        \hline
        Mean $\delta$ & .0274 \\
        \hline
        \% of ($\delta > 0$) & 78.0130 \\
        $P_{\chi} (\delta > 0)$ & .0000 \\
        \% of ($P_{\text{KW}} (\delta > 0) < .05$) & 22.8013 \\
        \hline
        Mean $\Delta_c$ & .2407 \\
        \hline
    \end{tabular}
    \vspace{0.3cm}
    \captionsetup{justification=centering}
    \caption*{\textit{Table 1: Activation vs. Dimensional Proximity Cross-analysis (Layer 1).}}

\end{table}

\noindent These results are consistent with our hypothesis: for a given neuron, the higher the activation of a token, the more it is categorically linked to the principal categorical sub-dimensions associated with that neuron. In other words, the magnitude of a token's activation is a positive function of the coordinates of that token within the categorical vector space constituted by the neuron's principal categorical sub-dimensions.

\subsection*{Relationship between Activation and Dimensional Proximity: Correlation}

We continue exploring our research question: is there a link between a token's activation value and the number of principal vectorial categorical sub-dimensions to which this token is associated? We now investigate the potential functional relationship between activation level and the number of categorical sub-dimensions involved.

More specifically, at the operational level, we proceeded as follows to complement our methodological approach compared to our previous investigation:
\begin{itemize}
    \item Since our data did not fully follow normal distributions, as confirmed by inferential tests (Shapiro-Wilk, Lilliefors, Kolmogorov-Smirnov, Jarque-Bera), we used Kendall's $\tau$, which allows us to investigate an ordinal relationship that does not strictly depend on linear and numerical conditions.
    \item We selected only the 612 neurons from layer 1 of GPT2-XL that were associated with exactly three principal categorical sub-dimensions, i.e., three taken-clusters, provided that each taken-cluster contained at least six tokens.
    \item For each of these target neurons, its 100 constitutive core-tokens were ranked by activation level, from lowest to highest. The classification of categorical sub-dimensions was performed in ascending order based on the number of tokens contained in their respective taken-clusters.
    \item An average neuron was then computed, where, for each token activation rank, its mean proximity score to each of the three sub-dimensions $D_1$, $D_2$, and $D_3$ was calculated. Kendall’s $\tau$ coefficients were determined on this average neuron.
\end{itemize}

Table n°2 shows very strong non-linear correlations, ranging from .78 to .85, between token activation level and dimensional proximity, i.e., between token activation rank and the number of categorical sub-dimensions involved. These correlations are highly significant at the inferential level ($p < .0001$).

\begin{table}[H]
    \centering
    \renewcommand{\arraystretch}{1.3}
    \begin{tabular}{|c|c|}
        \hline
        N & 612 \\
        \hline
        $\tau_{D_1}$ & .7754 \\
        \hline
        $p(\tau_{D_1})$ & .0000 \\
        \hline
        $\tau_{D_2}$ & .8384 \\
        \hline
        $p(\tau_{D_2})$ & .0000 \\
        \hline
        $\tau_{D_3}$ & .8461 \\
        \hline
        $p(\tau_{D_3})$ & .0000 \\
        \hline
    \end{tabular}
    \vspace{0.3cm}
    \captionsetup{justification=centering}
    \caption*{\textit{Table n°2: Ordinal correlation between mean activation and mean dimensional proximity (layer 1).}}
\end{table}

Graph n°1, produced from the same data, clearly shows the strong relationship linking activation level and dimensional proximity. For each of the three categorical sub-dimensions, we observe an almost linear relationship—even a slight exponential dynamic for the highest values—between token activation and their dimensional proximity to the sub-dimensions in question. Note that this relationship is very systematic, with extremely few, if any, outlier values that do not conform to this strongly manifested functional relationship.

\begin{figure}[H]
    \centering
    \makebox[\textwidth][l]{\hspace{-1cm} \includegraphics[width=1.2\textwidth]{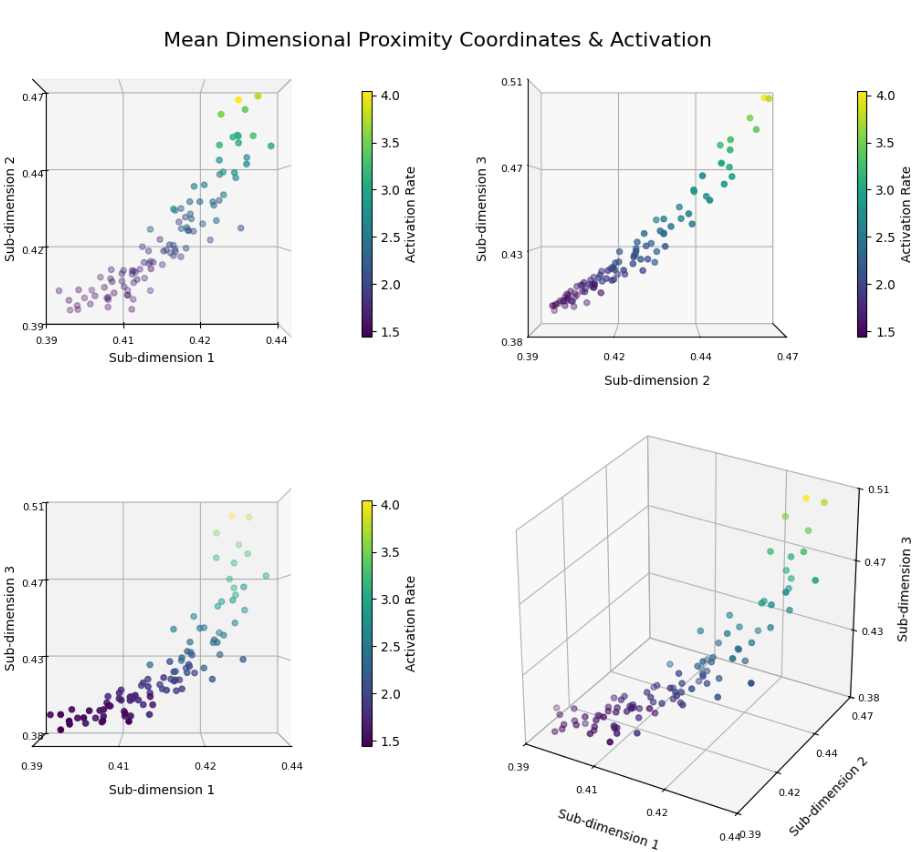}}
    \captionsetup{justification=centering, font=small}
    \caption*{\textit{Graph n°1 : Comparison of mean activations between categorical clusters from hierarchical classification on tokens' embeddings (layer 0).}}
\end{figure}

These data are again consistent with our hypothesis: at the neuronal level, the higher the activation level of a token, the more the neurons concerned are categorically linked to the principal categorical sub-dimensions associated with those neurons. There appears to be a strong functional relationship between activation level and sub-dimensional proximity.

\subsection{Structure of Categorical Sub-Dimensions}

To conclude, let us examine the structuring of categorical sub-dimensions. Are they subject to a particular organization among themselves? To what extent can we consider that they do not form orthogonal bases in terms of vector space? Is this organization related to the questions of categorical convergence and divergence of tokens within neurons, and therefore to the issue of neuronal polysemy?

To investigate this question, we proceeded methodologically as follows:
\begin{itemize}
    \item Within layer 1 of GPT2-XL, we selected the 463 neurons that: (i) were associated with exactly 4 (no more, no less) principal categorical sub-dimensions (i.e., taken-clusters), and (ii) had taken-clusters composed of at least 6 tokens.
    \item For each selected neuron, its 100 core-tokens were ranked in ascending order based on their activation level within that neuron.
    \item For each selected neuron, its 4 categorical sub-dimensions were ranked in descending order based on the number of tokens contained in their associated taken-cluster (thus, sub-dimension 1 is the one whose taken-cluster contains the most tokens).
    \item Each core-token rank (from 1 to 100) was averaged by calculating its mean categorical (i.e., dimensional) proximity with each of the 4 ordered categorical sub-dimensions, which were themselves averaged.
    \item This process resulted in the creation of a "mean neuron," where each of the 100 mean core-tokens was associated with 4 dimensional proximities, one for each averaged categorical sub-dimension.
    \item A principal component analysis (PCA) was then conducted, using the 100 mean core-tokens as statistical units and the 4 mean dimensional proximities related to the 4 categorical sub-dimensions as variables.
\end{itemize}

\begin{figure}[H]
    \centering
    \includegraphics[width=0.6\textwidth]{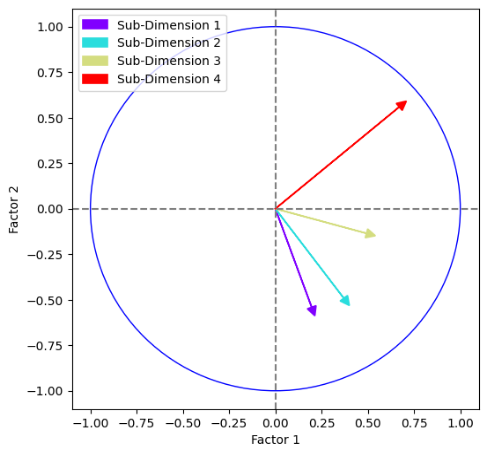}
    \captionsetup{justification=centering, font=small}
	\caption*{\textit{Graph n°2: Correlation circle (PCA) on the average neuron associated with layer 1.}}
\end{figure}

The graph n°2 presents the correlation circle we obtained. The PCA from which it was generated meets the fundamental statistical conditions of this factorial analysis procedure: (i) the sphericity of the variables ($P_{\text{Bartlett}} < .0001$) and (ii) the adequacy of the sampling ($KMO = .82 > .7$). The two main factors obtained are associated with eigenvalues $VP_1 = 3.74$ and $VP_2 = .19$, with the first axis, which is largely dominant, explaining 93.5\% of the initial variance, while the second accounts for 4.8\%. Several observations are particularly noteworthy:
\begin{itemize}
    \item The first factor, which is again the most dominant in terms of preserved variance, shows that all variables (the categorical sub-dimensions) are positively correlated with one another. This axis thus expresses partial categorical convergence (relative monosemy) between the different categorical sub-dimensions associated with a given neuron.
    \item The second factor contrasts certain categorical sub-dimensions (negative correlations). It highlights neuronal polysemy and categorical divergence, where certain tokens are preferentially more associated with some categorical sub-dimensions than with others. Given that this factor explains significantly less variance, this polysemy seems to have a weaker "effect size" (to use the term analogically) compared to the previously mentioned categorical convergence.
    \item Additionally, considering that sub-dimensions are ranked in descending order based on the number of tokens contained in their associated taken-cluster (sub-dimension 1 is the one with the largest taken-cluster, etc.), an interesting pattern emerges within this second factor. This second axis of neuronal polysemy appears to contrast sub-dimensions based on the number of tokens in their taken-clusters: taken-clusters with larger token counts are oriented towards the negative values of this factor, whereas the taken-cluster associated with sub-dimension 1 (in red on the graph) corresponds to a smaller number of tokens. This factor might suggest that categorical proximity between sub-dimensions is related to the number of tokens associated with their respective taken-clusters.
    \item Some categorical sub-dimensions are relatively orthogonal to one another (at least on this factorial plane): sub-dimension 4 stands out from all the others. Others are relatively collinear: sub-dimensions 1 and 2. Notably, factor 2 separates two groups of categorical sub-dimensions, one of which consists of convergent sub-dimensions with relatively low polysemy.
    \item The two axes together form a factorial plane that illustrates a partial categorical convergence, and thus a relative polysemy, among the categorical sub-dimensions involved.
\end{itemize}

Graph n°3 presents the correlation circles associated with different non-averaged neurons from layer 1. The graph confirms the main observational trends we have identified, as the averaging operation could have potentially produced the elements we indicated artificially.

\begin{figure}[H]
    \centering
    \makebox[\textwidth][l]{\hspace{-3cm} \includegraphics[width=1.6\textwidth]{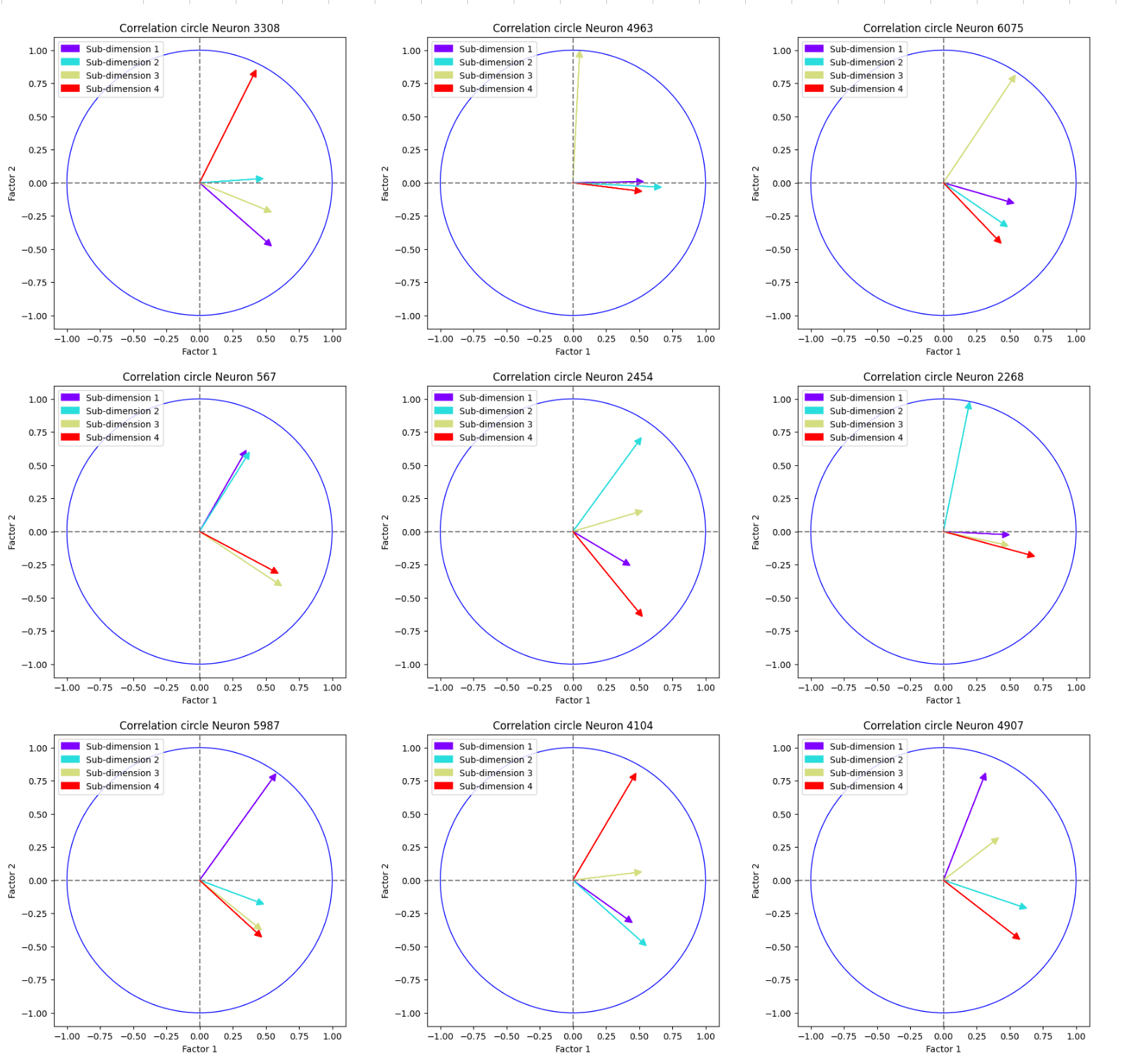}}
    \captionsetup{justification=centering, font=small}
    \caption*{\textit{Graph n°3: Correlation circles (PCA) for different neurons in layer 1.}}
\end{figure}

Graph n°4 allows us to observe the distribution of mean token activation levels on the factorial axes we obtained, relative to categorical sub-dimensions. We identify an important and very clear pattern that we previously highlighted: tokens with higher activation levels are those that simultaneously exhibit (i) the highest positive values on Factor 1 and (ii) the highest negative values on Factor 2. It is understood that the negative segment of this second factor, as we have already pointed out, is positively linked to the most significant dimensional proximity variables (i.e., those associated with taken-clusters containing the largest numbers of tokens) and the most numerous ones (3 out of 4 sub-dimensional variables). Once again, we observe that tokens with higher activation levels are those associated with the largest number of categorical sub-dimensions.

\begin{figure}[H]
    \centering
    \includegraphics[width=0.6\textwidth]{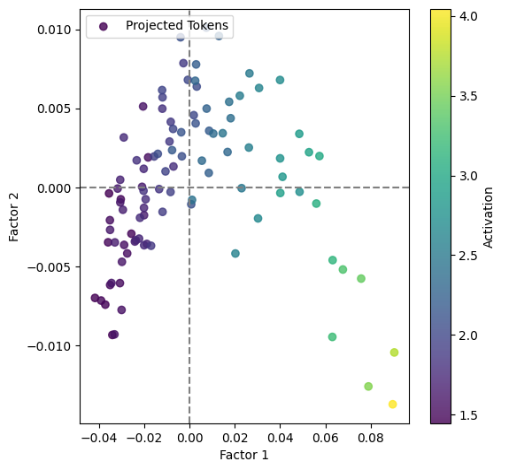}
    \captionsetup{justification=centering, font=small}
    \caption*{\textit{Graph n°4: Projection of tokens on factorial axes, with activations (layer 1).}}
\end{figure}

\section{Discussion of Results}

\subsection{Summary of Our Findings}

In this study, we investigated the hypothesis of a relationship between token activation levels and their coordinates within the vector space formed by the categorical sub-dimensions of the involved neurons. This was examined in the context of the first two perceptron layers of the GPT2-XL language model to evaluate the possibility of an alternative framework for interpreting and modeling synthetic neuronal polysemy.

By analyzing this relationship through the comparison of groups of core-tokens with very high or low activation levels, we observed that the higher the activation of a token, the more it is categorically linked to the principal categorical sub-dimensions of the involved neuron (and thus more monosemic). Examining this connection in terms of a functional relationship between token activation level and the number of categorical sub-dimensions concerned by this token, we found strong functional relations between activation intensity and sub-dimensional coordinates (i.e., proximities). Finally, investigating this relationship in the context of the structuring of categorical sub-dimensions, we observed that the highest activation values are associated with the highest coordinate values on the factorial axes reducing the categorical sub-dimensions (and thus with stronger monosemy). The first obtained factor expresses partial categorical convergence (relative monosemy) among the different categorical sub-dimensions linked to a given neuron, while the second denotes neuronal polysemy and categorical divergence.

These results are compatible with our hypothesis that the polysemic thought category of a neuron can be geometrically modeled as a categorical vector space whose basis consists of the categorical (monosemic) sub-dimensions extracted by the neuron from its preceding-layer neurons. This categorical vector space appears to be structured by the activation space of the neuron as follows: within this categorical vector space, highly activated core-tokens tend to simultaneously involve the various principal categorical sub-dimensions of this space, which leads to categorical convergence (monosemy) \cite{Pichat2024b} among these tokens. This is in contrast to less activated core-tokens, each of which tends to mobilize only a single categorical dimension, thereby provoking categorical divergence (polysemy) \cite{Pichat2024a} among these tokens. As previously mentioned, this divergence manifests in the form of categorical discontinuity and categorical inhomogeneity \cite{Pichat2024a}.

These categorical sub-dimensions, which constitute the type of vector space we hypothesize, result from the synthetic process of categorical clipping \cite{Pichat2024d}, through which artificial cognition extracts, at the level of an arrival neuron, categorical sub-dimensions from the thought categories of its precursor neurons. This clipping is itself driven by categorical priming, inter-neuronal categorical attention, and categorical phasing \cite{Pichat2024c}, guided by the neuronal aggregation function. These categorical sub-dimensions exhibit partial categorical confluence (relative polysemy) \cite{Pichat2025} and thus determine a non-orthogonal basis, though one whose vectors remain linearly independent.

These sub-dimensions are associated with activation dispersion \cite{Pichat2025}, meaning that they do not correspond to continuous segments of token activation values within the activation space of either their source or destination neurons. Instead, activation space influences the generation of categorical sub-dimensions in an indirect manner: through intra-neuronal attention \cite{Pichat2025a}, which allows a neuron to locate and isolate a relatively homogeneous categorical segment—namely, a critical concept-in-act (monosemic) \cite{Vergnaud2009, Vergnaud2016}—within the polysemic thought category it carries. Then, based on these synthetic paroxysmal concepts-in-act, which serve as the source of categorical priming, categorical sub-dimensions (monosemic) are clipped, reinforced by activation through inter-neuronal attention and categorical phasing. The latter effect, categorical phasing, plays a central role in our study, as it fundamentally accounts for the fact that the most activated tokens within the semantic vector space we postulate are those that simultaneously belong to multiple categorical sub-dimensions constitutive of this vector space, and therefore tend to be monosemic (at least with respect to the conceptual criterion constructed by the involved neuron).

\subsection{Interpretation Differences in Polysemy}

Previous research on neuronal polysemy \cite{Bricken2023, Haider2025, Nanda2023} hypothesizes that it plays a crucial role in encoding more categories of thought than there are neurons available in a Multi-Layer Perceptron (MLP) network layer. In this framework, a synthetic category is represented through the coordinated activation of multiple neurons, allowing the same neurons to contribute to various categories of thought. This proposed neuronal strategy enhances the network’s capacity to represent a greater number of potential synthetic categories. Such a mechanism is believed to improve the efficiency of large natural language processing models by leveraging the principle of superposition, where a single resource can be used to represent multiple pieces of information.

These studies have generally conceptualized neuronal polysemy as a process that emerges at the level of neuronal activations within a vector space composed of the outputs of numerous neurons in a given layer. However, we propose an alternative, non-exclusive, perspective on this idea. Rather than focusing on polysemy through an interpretative approach centered on neurons within a layer, we suggest that the source of this polysemy could also be conceived as residing within each neuron itself, with the neuron capable of containing multiple categorical sub-dimensions. In other words, each neuron could itself be a complex entity capable of hosting different intra-neuronal facets of meaning, rather than these being distributed solely in an inter-neuronal manner.

Our perspective does not seek to contradict existing theories emerging from the interaction of multiple neurons but rather offers an alternative interpretative approach. Specifically, we propose that each neuron can be understood as internally constructing an independent categorical vector space (separate from its parallel neurons in the same layer), with its basis formed by categorical sub-dimensions extracted, via categorical clipping, from preceding categorical dimensions. Within this space, each token is represented as a point whose coordinates are proportional to the semantic proximity linking it to the involved categorical sub-dimensions. 

In this space, each token with moderate or low activation falls into a polysemic zone, which is relatively low in categorical homogeneity. Within this zone, the activation space does not necessarily delineate closely related categorical segments, as two neighboring activations may correspond to different categorical sub-dimensions. In contrast, a singular zone of this space, associated with very high activation levels, defines a categorical node that appears more monosemic—a locus of categorical convergence. This is the intersection of very high values across each of the sub-dimensions at play, forming a focal point of intra-neuronal attention on the synthetic critical concept-in-act that the neuron seeks to detect, allowing for the subsequent extraction of a new, even more refined categorical sub-dimension.

\section{Conclusion}

Analyzing the knowledge contained within the perceptron-type synthetic neurons of language models, within an observational framework based on vector spaces composed of clipped categorical sub-dimensions, could serve as a relevant methodological approach for conceptualizing and operationalizing a genuine cognitive entanglement between synthetic and symbolic knowledge. Such a cognitive entanglement, at an extremely fine cognitive granularity, would enable the development of hybrid architectures that do not merely define the neuro-symbolic articulation in terms of post-processing activities, where a module of one type (e.g., neural) is processed by a module of another type (e.g., symbolic), or vice versa \cite{Sun2024}. 

By embedding itself within the paradigm of Vygotsky \cite{Vygotsky1934}, such an approach could facilitate the creation of composite, neuro-symbolic artificial intelligences, finely interwoven between "practical" synthetic neural knowledge—implicit, contextualized, and acquired through the learning phase of networks—and "scientific" human symbolic knowledge. The former would serve to ground and contextualize the latter, while the latter would function to systematize, generalize, formalize, explicate, and regulate the former.

This neuro-symbolic interfacing is likely to provide a particularly fruitful methodological framework for integrating a range of operational developments that contemporary artificial intelligence architectures could benefit from, as proposed by cognitive sciences and human neuroscience \cite{Minsky1988, Sun2024, Xie2023, Marcus2021}, as well as philosophical reflections on AI \cite{Efimov2021, Efimov2023}. This could support the evolution of artificial general intelligence toward more efficient, more intelligent, and safer systems, complementing the prevailing "more data" approaches that dominate current methodologies.

\section*{Acknowledgments}

The authors thank Madeleine Pichat for her careful proofreading of this article.

\end{document}